\documentclass[letterpaper]{article} 
\usepackage{aaai25}  
\usepackage{times}  
\usepackage{helvet}  
\usepackage{courier}  
\usepackage[hyphens]{url}  
\usepackage{graphicx} 
\usepackage{natbib}  
\usepackage{caption} 
\usepackage{algorithm}
\usepackage{algorithmic}

\usepackage{newfloat}
\usepackage{listings}
\DeclareCaptionStyle{ruled}{labelfont=normalfont,labelsep=colon,strut=off} 
\floatstyle{ruled}
\newfloat{listing}{tb}{lst}{}
\floatname{listing}{Listing}
\usepackage{amsmath} 
\usepackage{amssymb} 
\usepackage{booktabs} 
\usepackage{multirow} 
\usepackage{colortbl} 
\usepackage{subcaption} 
\usepackage{pdfpages}

\title{Efficient Few-Shot Neural Architecture Search\\by Counting the Number of Nonlinear Functions}
\author{
    Youngmin  Oh\textsuperscript{\rm 1}, Hyunju Lee\textsuperscript{\rm 1}, and Bumsub Ham\textsuperscript{\rm 1,2}\thanks{Corresponding author.}
}
\affiliations{
    \textsuperscript{\rm 1}Yonsei University~~~~\textsuperscript{\rm 2}Korea Institute of Science and Technology~(KIST)

    \{youngmin.oh,~hyunjulee,~bumsub.ham\}@yonsei.ac.kr
}

\usepackage{bibentry}

\begin{document}

\maketitle

\begin{abstract}
Neural architecture search (NAS) enables finding the best-performing architecture from a search space automatically. Most NAS methods exploit an over-parameterized network (\emph{i.e.}, a supernet) containing all possible architectures (\emph{i.e.}, subnets) in the search space. However, the subnets that share the same set of parameters are likely to have different characteristics, interfering with each other during training. To address this, few-shot NAS methods have been proposed that divide the space into a few subspaces and employ a separate supernet for each subspace to limit the extent of weight sharing. They achieve state-of-the-art performance, but the computational cost increases accordingly. We introduce in this paper a novel few-shot NAS method that exploits the number of nonlinear functions to split the search space. To be specific, our method divides the space such that each subspace consists of subnets with the same number of nonlinear functions. Our splitting criterion is efficient, since it does not require comparing gradients of a supernet to split the space. In addition, we have found that dividing the space allows us to reduce the channel dimensions required for each supernet, which enables training multiple supernets in an efficient manner. We also introduce a supernet-balanced sampling (SBS) technique, sampling several subnets at each training step, to train different supernets evenly within a limited number of training steps. Extensive experiments on standard NAS benchmarks demonstrate the effectiveness of our approach. Our code is available at \url{https://cvlab.yonsei.ac.kr/projects/EFS-NAS}.
\end{abstract}
\vspace{-.35cm}

%

\begin{figure}[t]
	\small
	\centering
    \includegraphics[width=\linewidth]{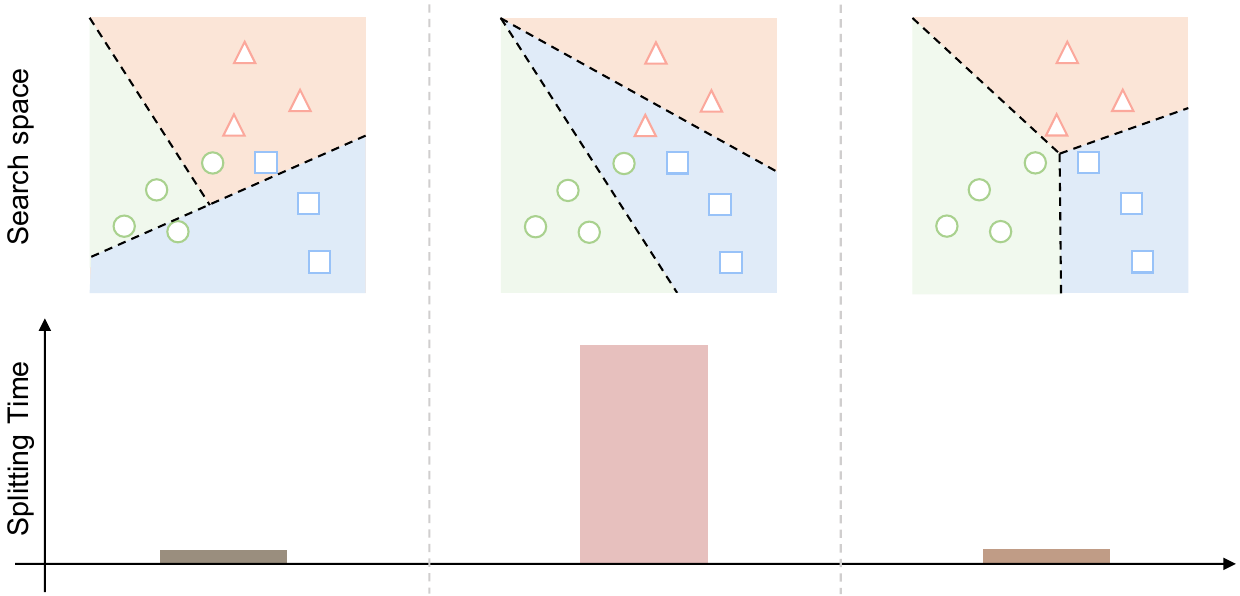}
    \vspace{-.6cm}
    \caption{Illustration of search space splitting strategies. Individual supernets are highlighted in different colors. Subnets with similar characteristics are marked by the same shape. \textbf{Left:} FS-NAS~\cite{zhao2021few} splits the space randomly. Although the random splitting strategy is efficient, each supernet could contain subnets that are likely to conflict with each other. \textbf{Middle:} GM-NAS~\cite{hu2022generalizing} compares gradients of a supernet to split the space, better grouping subnets. This however incurs a lot of computational cost. \textbf{Right:} We propose to count the number of nonlinear functions within a subnet such that each subspace contains subnets with the same number of nonlinear functions only. Our splitting criterion incurs negligible overheads, while separating the space effectively.
    Best viewed in color.}
    \vspace{-.8cm}
    \label{fig:fig1}
\end{figure}
\section{Introduction}
Manually designing network architectures is a labor-intensive and time-consuming process. Neural architecture search (NAS) helps to automate the designing process, and provides optimal network architectures for various hardware configurations~(\emph{e.g.}, FLOPs). Early NAS methods~\cite{zoph2018learning,baker2016designing,zoph2016neural} adopt reinforcement learning~\cite{williams1992simple} with policy networks~(\emph{i.e.}, controllers), which requires training deep neural networks from scratch, taking lots of computational costs~(\emph{e.g.}, typically thousands of GPU hours). To overcome this problem, one-shot NAS approaches~\cite{liu2018darts,guo2020single,xu2019pc,chu2021fairnas,cai2018proxylessnas} adopt a weight-sharing technique~\cite{pham2018efficient}, where they train a single supernet that consists of all possible network architectures~(\emph{i.e.}, subnets) in a given search space. The trained supernet can act as a performance estimator for various subnets, sampled from the supernet, indicating that each subnet does not need to be trained from scratch to predict its performance. Although one-shot NAS methods are efficient, they are limited in that the performance of the subnet estimated from the supernet is less correlated with the one obtained from training the subnet from scratch. The major reason is that the parameters of the supernet suffer from conflicts between subnets during training. To address this problem, few-shot NAS methods~\cite{zhao2021few,hu2022generalizing,su2021k} propose to exploit multiple supernets. They typically divide a search space into multiple subspaces, and then train an individual supernet for each subspace. Since each supernet has its own parameters, subnets from different supernets do not interfere with each other during training. This allows the estimated performance of subnets to become more correlated with the actual one, which however requires at least a few times more computational cost than one-shot methods.

We introduce in this paper a novel few-shot NAS method that splits the search space in an efficient manner, while better alleviating the conflicts between subnets. Specifically, we leverage the number of nonlinear functions~(\emph{e.g.}, ReLU~\cite{krizhevsky2012imagenet}) within a particular subnet to split the search space such that each subspace contains subnets with the same number of nonlinear functions only. We then assign a separate supernet to each subspace to prevent subnets from different supernets from interfering with each other. In particular, our method results in subnets from the same supernet having similar characteristics in terms of the number of parameters, FLOPs, and test accuracy, making them less likely to suffer from the conflicts. Different from current few-shot NAS methods~(See Fig.~\ref{fig:fig1}), our splitting criterion is more efficient, since it does not require comparing gradients of a supernet to split the search space~\cite{hu2022generalizing} and a specific technique for initializing supernets~\cite{zhao2021few,hu2022generalizing}. In addition, we have found that effectively dividing the search space helps to maintain the performance ranking between subnets, even when the number of channels varies. Based on this observation, we propose to adjust the number of channels for each supernet to further improve the efficiency, allowing us to train all supernets on a single machine in contrast to existing few-shot methods. We also introduce a supernet-balanced sampling~(SBS) technique for better training supernets. Since each supernet has a different number of subnets, randomly sampling one of the subnets at each training step would be biased towards training the supernet with the largest number of subnets. Instead of selecting a single subnet at each training step, our SBS samples multiple subnets from different supernets. This enables training different supernets evenly within a limited number of training steps. Extensive experiments on NAS201~\cite{dong2020bench} and ImageNet~\cite{deng2009imagenet} demonstrate that our approach achieves state-of-the-art results with much less computational overheads than other few-shot methods~\cite{zhao2021few,hu2022generalizing,su2021k}. We summarize our main contributions as follows:
\begin{itemize}
    \item[$\bullet$] We introduce a simple yet effective method using the number of nonlinear functions to split the space. Our method enables reducing the number of channels for each supernet, providing much smaller computational cost, compared with current few-shot methods.
    \item[$\bullet$] We present a SBS scheme, where multiple subnets are sampled from different supernets at each training step, to train different supernets equally.
    \item[$\bullet$] We demonstrate the effectiveness of our approach on NAS201~\cite{dong2020bench} and ImageNet~\cite{deng2009imagenet}, and provide extensive experiments along with ablation studies.
\end{itemize}


\section{Related work}
NAS aims to find a well-performing network in a search space efficiently. To this end, the search space, typically defined as the total number of layers and a set of candidate operations~(\emph{e.g.}, convolutional or pooling layers), should cover various network architectures~\cite{zhou2021autospace,ci2021evolving,radosavovic2020designing,yu2020bignas}, and the searching process should be effective and efficient~\cite{guo2020single,you2020greedynas,chu2021fairnas}. In the following, we describe representative methods pertinent to ours.

\noindent \textbf{One-Shot NAS.} Many NAS methods adopt a weight-sharing strategy that trains an over-parameterized network~(\emph{i.e.}, a supernet) containing all possible operations at each layer in order to reduce the search time. The seminal work of~\cite{liu2018darts} proposes to compute a weighted average of feature maps at each layer, where each feature map is obtained from a corresponding operation. It thus needs to train all operations at training time, resulting in a large memory footprint. To reduce the memory requirement, several methods~\cite{guo2020single,pham2018efficient,you2020greedynas,lu2023pa} propose to sample and train a single operation at each layer only~(\emph{i.e.}, a subnet). For example, SPOS~\cite{guo2020single} uniformly samples one of the subnets at each training step and updates corresponding operations only, reducing the computational cost. Instead of equally treating all subnets, GreedyNAS~\cite{you2020greedynas} focuses more on sampling subnets that are likely to perform better than others, but this requires evaluating a set of candidate subnets on a validation set repeatedly. On the other hand, FairNAS~\cite{chu2021fairnas} attempts to sample every operation at each layer more evenly. To this end, it selects several subnets at each training step and updates entire operations for the subnets simultaneously. Our SBS is similar to FairNAS in that it samples multiple subnets at each training step. On the contrary, SBS aims at training different supernets evenly, preventing the supernet with the largest number of subnets from being sampled more during training. Specifically, SBS forces subnets to be sampled from different supernets, while the subnets sampled from FairNAS belong to the same supernet.

\noindent \textbf{Few-Shot NAS.} A few methods have been introduced to use multiple supernets in NAS, which can be divided into two groups depending on whether they split a search space or not. First, recent methods~\cite{zhao2021few,hu2022generalizing} propose to split a search space into a set of subspaces and then train a separate supernet for each subspace. For example, GM-NAS~\cite{hu2022generalizing} formulates splitting the search space as a graph clustering problem. To be specific, it first computes gradients for individual operations at each layer, and measures cosine similarity between all pairs of the gradients. It then applies the graph min-cut algorithm~\cite{stoer1997simple} with setting the cosine similarity as a cut cost, and splits the search space so that subnets from the same subspace are less likely to interfere with each other. GM-NAS however requires training a single supernet that covers the entire search space to initialize subsequent supernets, incurring extra overheads. Additionally, it trains a set of supernets in a sequential manner, due to the large computational cost. Our approach also divides the search space into a set of subspaces, but differs in that (1) it does not train the single supernet covering the entire search space for initialization and (2) it allows to train all supernets simultaneously. Second, $K$-shot NAS~\cite{su2021k} proposes to duplicate parameters of each operation $K$ times~(\emph{i.e.}, $K$ copies of the supernet) without splitting the search space. In particular, it produces a subnet by computing a weighted average of the duplicated parameters, rather than sampling a single operation at each layer as in other methods~\cite{guo2020single,zhao2021few,hu2022generalizing}. Specifically, a generator is trained along with the copies of the supernet to produce $K$-dimensional probabilities for the weighted average. Similar to ours, $K$-shot NAS trains supernets simultaneously and does not require a specific scheme for initializing supernets. Although all the aforementioned methods for few-shot NAS alleviate the conflicts between subnets at training time, they are computationally expensive compared to one-shot approaches. Differently, based on our finding that effectively dividing the search space enables preserving the performance ranking between subnets, our approach reduces the number of channels for individual supernets to train multiple supernets efficiently. Note that this is different from early methods~\cite{liu2018darts,xu2019pc} that simply use a smaller supernet  due to their unacceptable overhead.

\noindent \textbf{Zero-Shot NAS.} Several methods have been introduced to avoid training supernets. They rely on training-free measurements, typically referred to as zero-cost proxies, to evaluate the performance of each subnet. For example, inspired by neural tangent kernels (NTKs)~\cite{jacot2018neural,lee2019wide} representing training dynamics of a neural network, the works of~\cite{chen2021neural,mok2022demystifying,xu2021knas} employ the condition number of a NTK for each subnet as its trainability. However, calculating a NTK for each subnet is computationally demanding. Another line of work~\cite{mellor2021neural,chen2021neural} instead uses the number of linear regions divided by ReLU activations during the forward pass of a network to measure the representational power of the network~\cite{hanin2019complexity,hanin2019deep}. While computing the number of linear regions is efficient in that it requires a single forward pass only, it cannot be applied for the case that neural networks adopt other activations (\emph{e.g.}, a tanh function). Recently, AZ-NAS~\cite{lee2024az} introduces a new measurement called an isotropy of a feature space. As the isotropy is obtained by computing similarities between intermediate features, it is applicable regardless of activation functions. Similar to AZ-NAS, our approach could be applicable to various activation functions, since it needs to count the number of nonlinear functions only. We differ from all the aforementioned methods in that our approach to counting the number of nonlinear functions focuses on dividing the search space for few-shot NAS rather than accurately measuring the performance of each subnet. In addition, our splitting criterion does not require processing forward and backward passes, which are computationally expensive for splitting the search space.


\section{Method}
In this section, we describe a weight-sharing technique for NAS briefly, and introduce our approach to using the number of nonlinear functions within a network to divide a search space. We then describe a SBS technique and how to sample optimal network architectures.

\subsection{Problem statement}
Let us suppose a search space~$\mathcal{A}$ consisting of subnets~$a_n$ as follows:
\begin{equation}
	\mathcal{A} = \{a_n \mid n=1,2,\dots,N\},
\end{equation}
where $N$ is the total number of subnets. To avoid training individual subnets from scratch, we adopt a weight-sharing technique that employs an over-parameterized network (\emph{i.e.}, a supernet) containing all candidate operations at each layer. The parameters of each subnet~$a_n$ are determined by selecting a corresponding operation at each layer of the supernet. Formally, we define the learnable parameters of the~$i$-th operation at the~$j$-th layer as follows:
\begin{equation}
	w(i,j) \in \mathbb{R}^{C_{\text{out}}(i,j) \times C_{\text{in}}(i,j) \times s(i,j) \times s(i,j)},
\end{equation}
where $C_{\text{out}}$ and $C_{\text{in}}$ are the numbers of output and input channels, respectively, and $s$ is the size of the filter. The parameters of the supernet for the entire search space~$\mathcal{A}$ can then be defined as follows:
\begin{equation}
	\mathcal{W}(\mathcal{A}) = \bigcup_{i,j} w(i,j).
\end{equation}
The weight-sharing scheme reduces the computational cost significantly, since we train the single supernet only. It however suffers from conflicts between subnets at training time, as all architectures in the search space share the same set of parameters $\mathcal{W}(\mathcal{A})$. To reduce the conflicts between subnets, few-shot NAS methods~\cite{zhao2021few,hu2022generalizing} propose to limit the extent of weight sharing by splitting the search space into subspaces and assigning an individual supernet to each subspace. Namely, subnets from different supernets do not share parameters. It is thus important to partition the search space effectively to minimize the interference between subnets sharing the same supernet. In the following, we describe our approach to splitting the search space in detail. 

\begin{figure*}[t]
	\small
    \centering
	\includegraphics[width=0.19\linewidth]{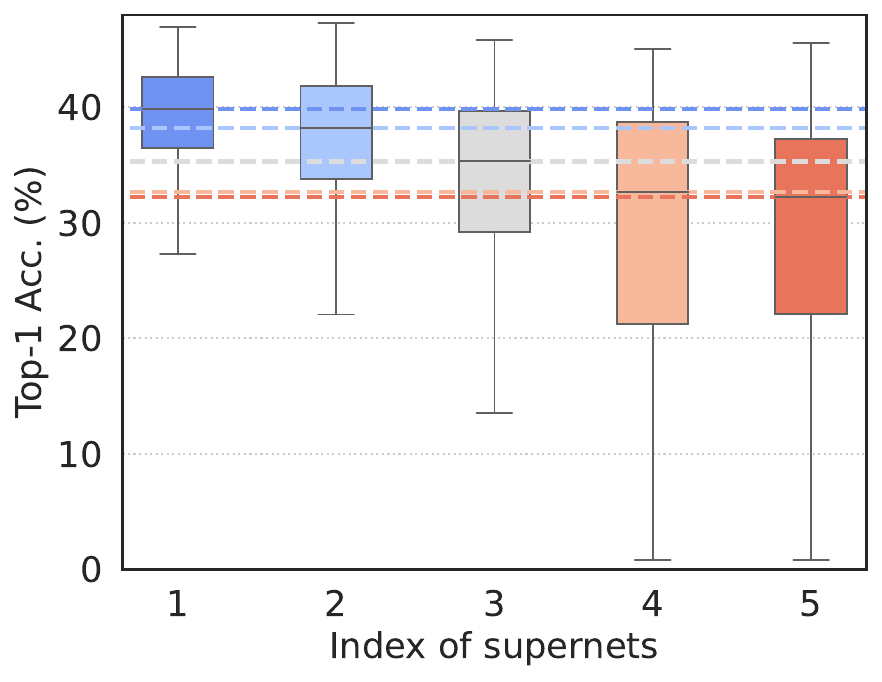}
	\includegraphics[width=0.19\linewidth]{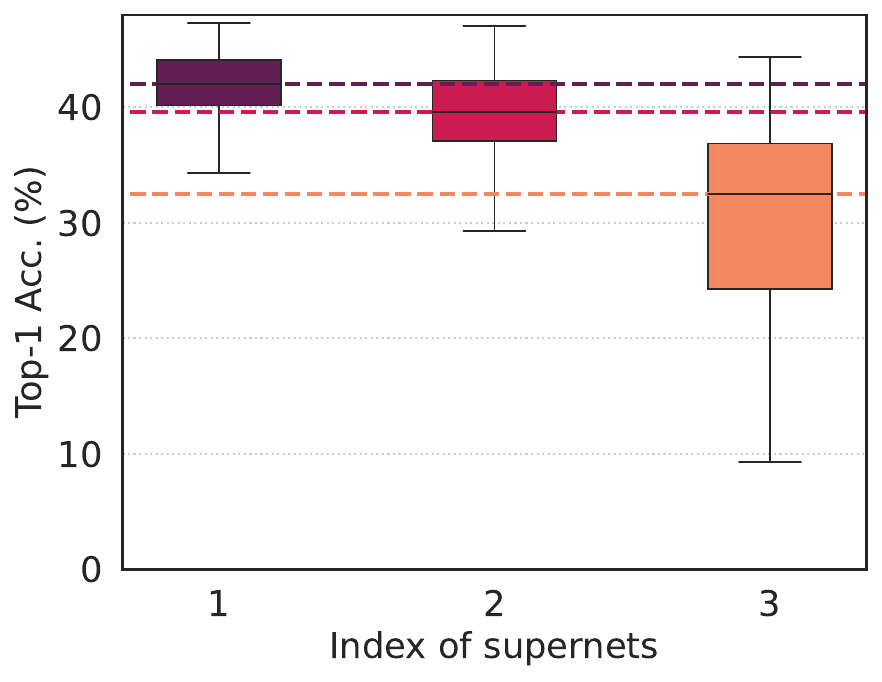}
	\includegraphics[width=0.19\linewidth]{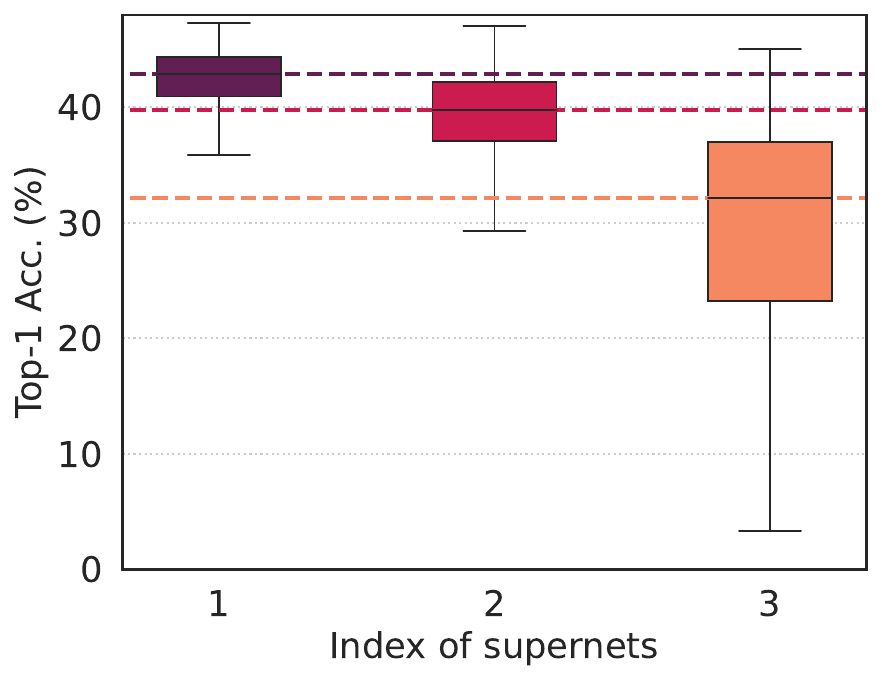}
	\includegraphics[width=0.19\linewidth]{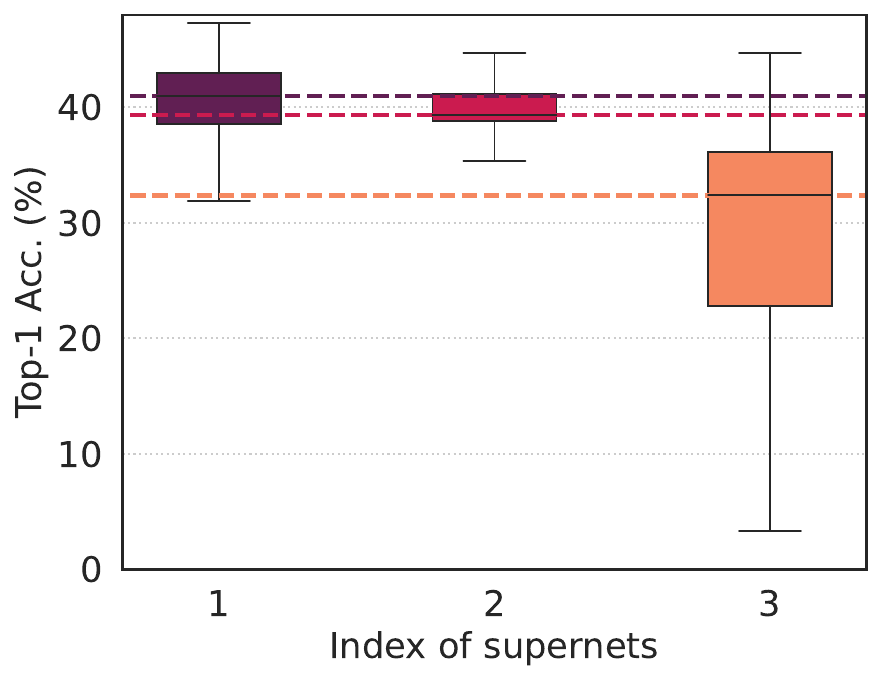}
	\includegraphics[width=0.19\linewidth]{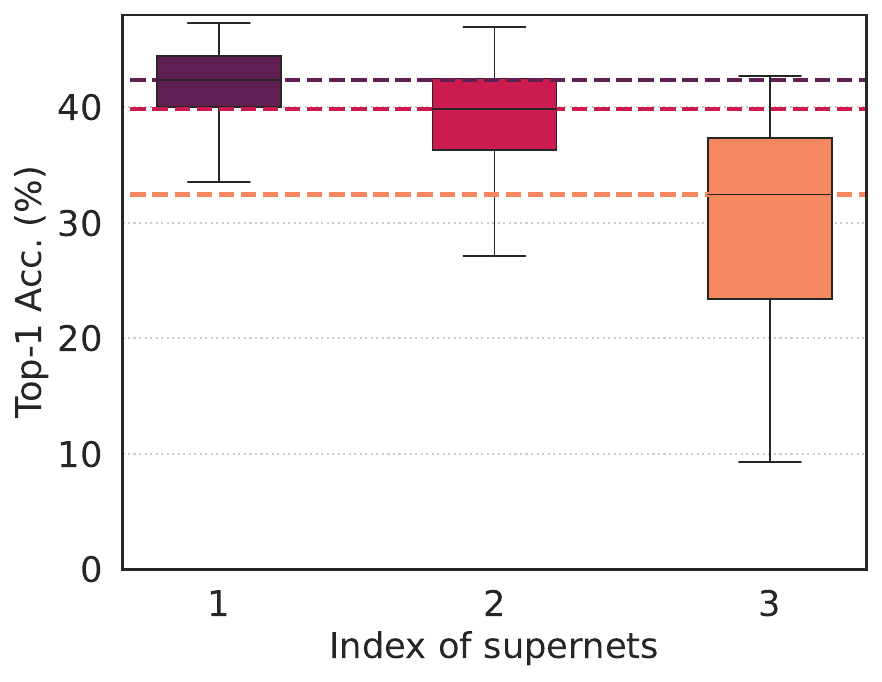}
    
    \begin{minipage}{0.19\linewidth}
    	\small{~~~~~~~~~~~~~~~~~~~~~(a)}
    \end{minipage}
    \begin{minipage}{0.19\linewidth}
    	\small{~~~~~~~~~~~~~~~~~~~~~(b)}
    \end{minipage}
    \begin{minipage}{0.19\linewidth}
    	\small{~~~~~~~~~~~~~~~~~~~~~(c)}
    \end{minipage}
    \begin{minipage}{0.19\linewidth}
    	\small{~~~~~~~~~~~~~~~~~~~~~(d)}
    \end{minipage}
    \begin{minipage}{0.19\linewidth}
    	\small{~~~~~~~~~~~~~~~~~~~~~(e)}
    \end{minipage}
    
    \vspace{-.2cm}
    \caption{Distributions of top-1 test accuracies for subnets of each supernet on ImageNet-16-120~\cite{chrabaszcz2017downsampled} of NAS201~\cite{dong2020bench}. Each dotted line represents the median value for the corresponding distribution. (a) FS-NAS~\cite{zhao2021few} adopts five supernets, dividing the search space randomly. (b-e) We leverage zero-cost proxies to divide the space, resulting in three supernets. They are (from left to right) FLOPs~\cite{ning2021evaluating,li2023zico}, the number of linear regions~\cite{mellor2021neural}, an isotropy of a feature space~\cite{lee2024az}, and the number of nonlinear functions. Best viewed in color.}
    \label{fig:fig2}
    \vspace{-.5cm}
\end{figure*}

\begin{figure*}[t]
	\small
	\centering
	\includegraphics[width=0.28\linewidth]{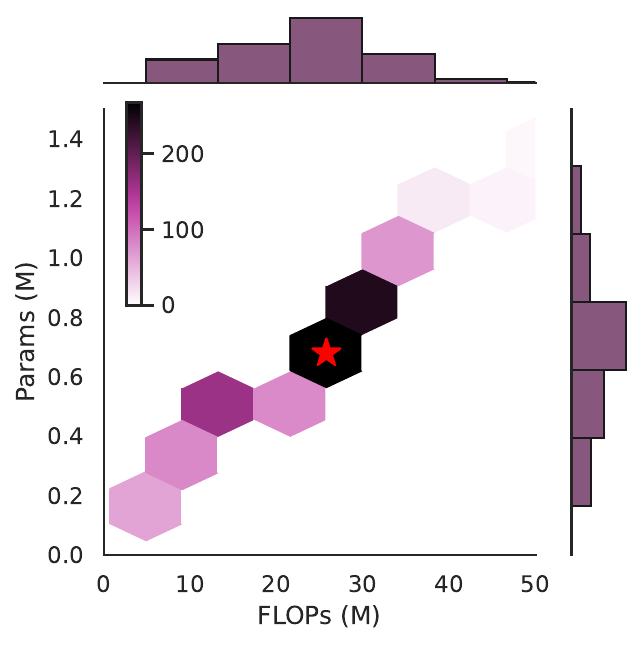}
	\includegraphics[width=0.28\linewidth]{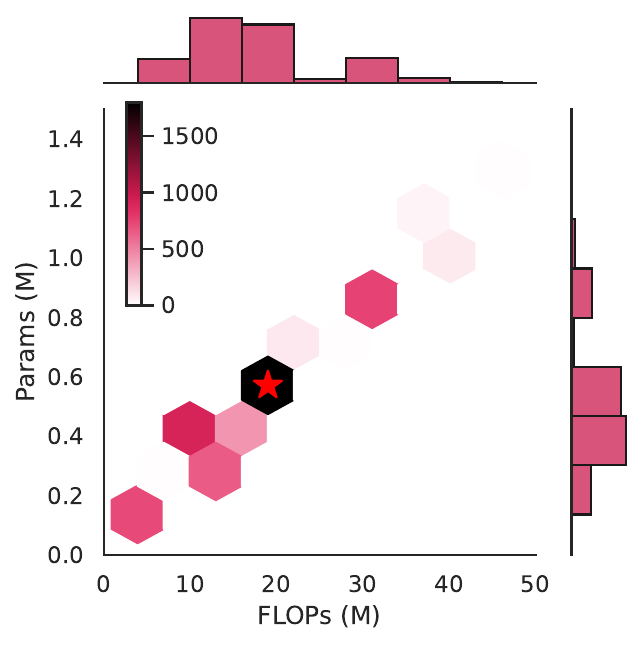}
	\includegraphics[width=0.28\linewidth]{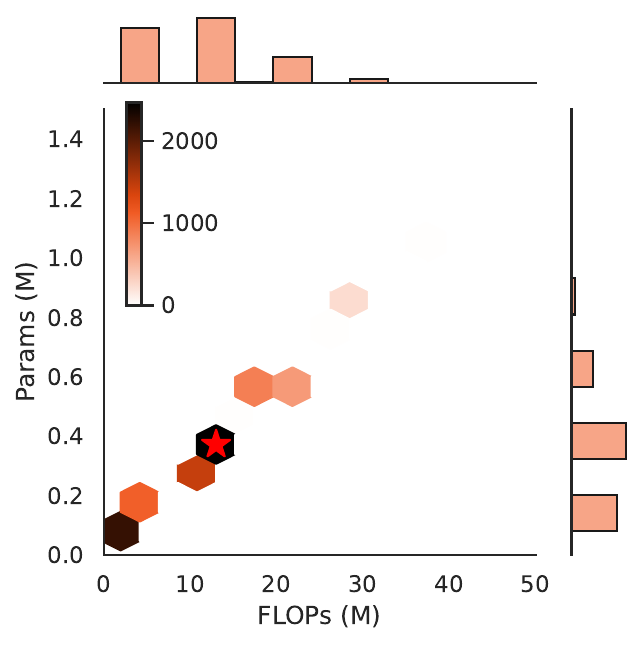}
    \vspace{-.4cm}
	\caption{Histograms of two variables (\emph{i.e.}, FLOPs and the number of parameters) for subnets of each supernet, where a red star indicates a bin with the highest frequency. We can see that our splitting criterion makes subnets from different supernets have unique structures in terms of FLOPs and the number of parameters. This suggests that our approach enables better training subnets, since the subnets with similar structures are less likely to suffer from the conflicts, compared with the ones with different structures. Best viewed in color.}
	\label{fig:fig3}
    \vspace{-.5cm}
\end{figure*}

\subsection{Division using the number of nonlinear functions}
\noindent \textbf{Analysis of zero-cost proxies.} To verify using the number of nonlinear functions as a splitting criterion, we present in Figs.~\ref{fig:fig2} and~\ref{fig:fig3} a statistical analysis of subnets on NAS201~\cite{dong2020bench} that provides stand-alone accuracies of the subnets computed by training them from scratch. Specifically, we explore four zero-cost proxies to divide a search space into a set of disjoint subspaces: FLOPs~\cite{ning2021evaluating,li2023zico}, the number of linear regions~\cite{mellor2021neural}, an isotropy of a feature space~\cite{lee2024az}, and the number of nonlinear functions. The proxy-based criteria provide three supernets on NAS201, while FS-NAS~\cite{zhao2021few} using a random splitting criterion employ five supernets. We can see in Fig.~\ref{fig:fig2} that the criteria using the zero-cost proxies show a clear difference between the supernets in terms of the median accuracy of the subnets, compared to the random splitting strategy in FS-NAS. This implies that the proxy-based criteria split the space effectively. In particular, it is worth noting that the criteria using other zero-cost proxies (\emph{i.e.}, FLOPs, the number of linear regions, and the feature isotropy) require a single forward pass for each subnet, while our criterion does not need the computation of forward passes. Considering that the number of subnets is typically innumerable, our criterion is more suitable for splitting the space. In addition, Fig.~\ref{fig:fig3} shows that our splitting criterion allows subnets from different supernets to have unique structures in terms of FLOPs and the number of parameters. This helps to mitigate conflicts between the subnets belonging to the same supernet, as subnets with similar structures are less likely to interfere with each other than those with different structures. Please refer to Sec.~3 in the supplementary material for a more detailed analysis.

\noindent \textbf{Supernets and subspaces.} Here we introduce a simple yet effective criterion that splits the search space into a set of disjoint subspaces. Concretely, with a function~$D(\cdot)$ counting the number of nonlinear functions (\emph{e.g.}, ReLU~\cite{krizhevsky2012imagenet}) within a subnet, we set a subspace whose subnets have the same number of nonlinear functions,~\emph{i.e.},~$k$, as follows:
\begin{equation}
	\mathcal{A}_k = \{a_n \mid D(a_n) = k~\text{and}~n=1,2,\dots,N_k\},
\end{equation}
where $\mathcal{A} = \bigcup_{k} \mathcal{A}_k$ and $N = \sum_{k} N_k$. We then assign a separate set of parameters for each subspace as follows:
\begin{equation}
	\mathcal{W}({\mathcal{A}_k}) = \bigcup_{i,j} w_k(i,j),
\end{equation}
where we denote by $w_k(i,j)$ the parameters of the $i$-th operation at the $j$-th layer for the supernet covering the subspace~$\mathcal{A}_k$.
In this way, we can prevent subnets from different subspaces from interfering with each other, allowing us to better train the subnets.
\begin{figure*}[t]
	\small
    \centering
    \includegraphics[width=0.19\linewidth]{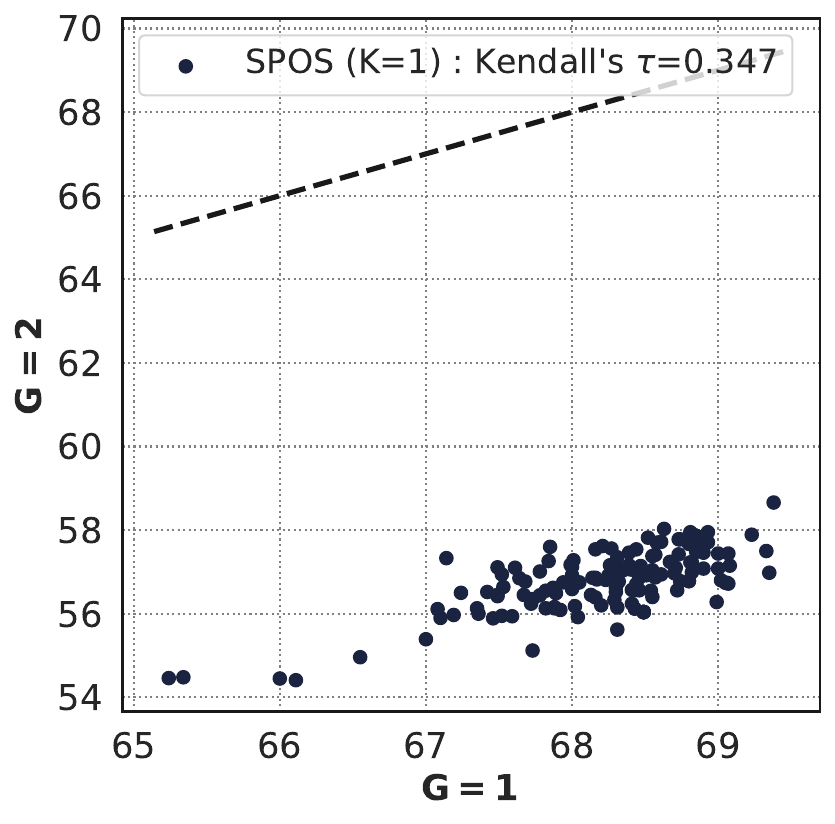}
    \includegraphics[width=0.19\linewidth]{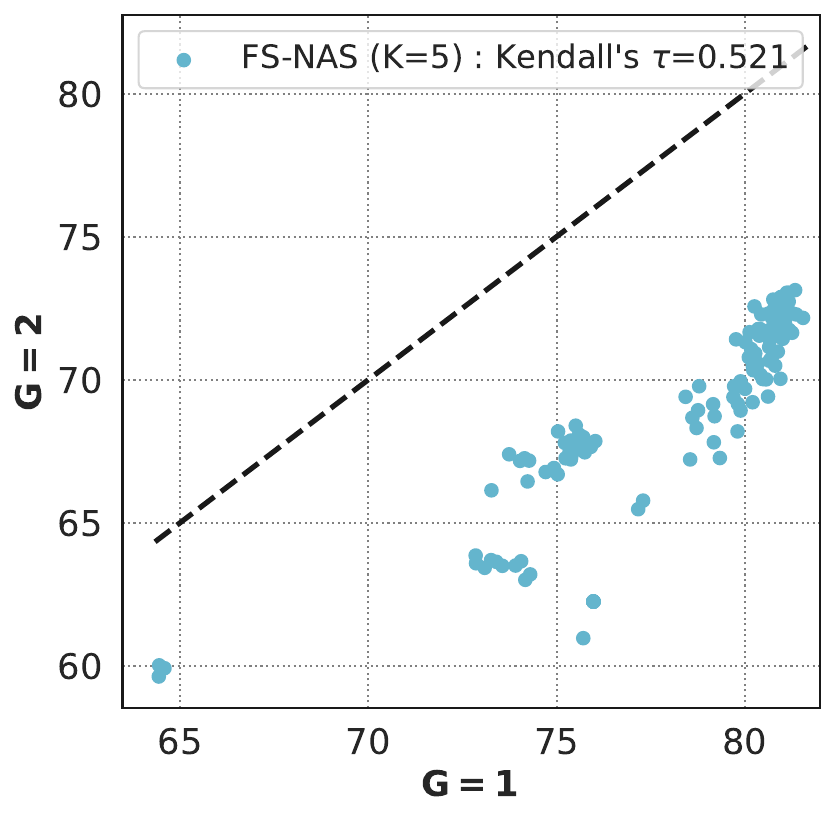}
    \includegraphics[width=0.19\linewidth]{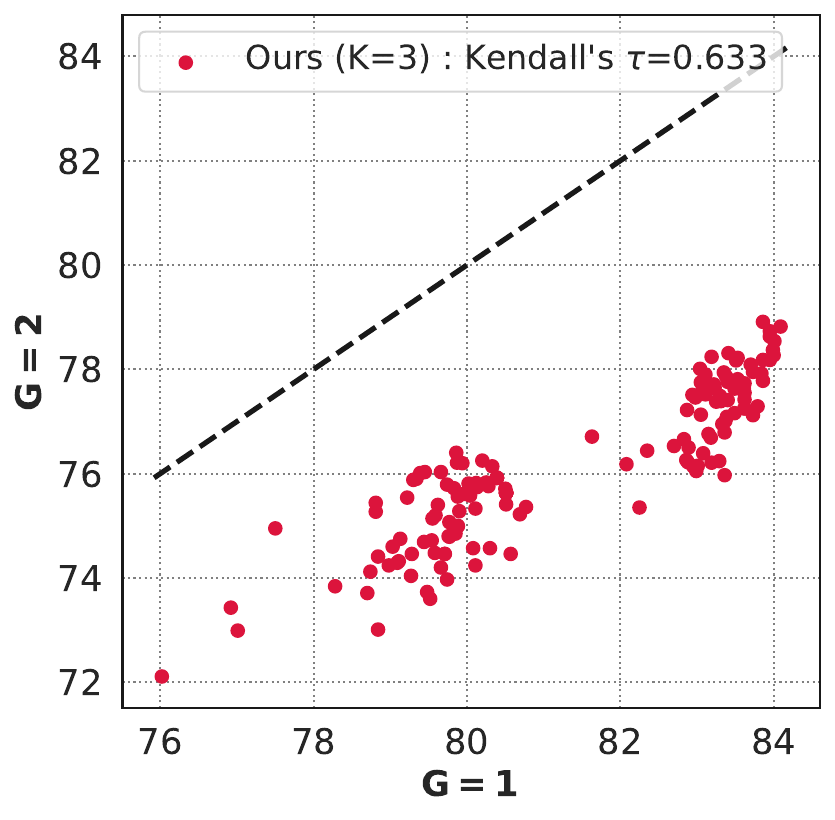}
    \includegraphics[width=0.19\linewidth]{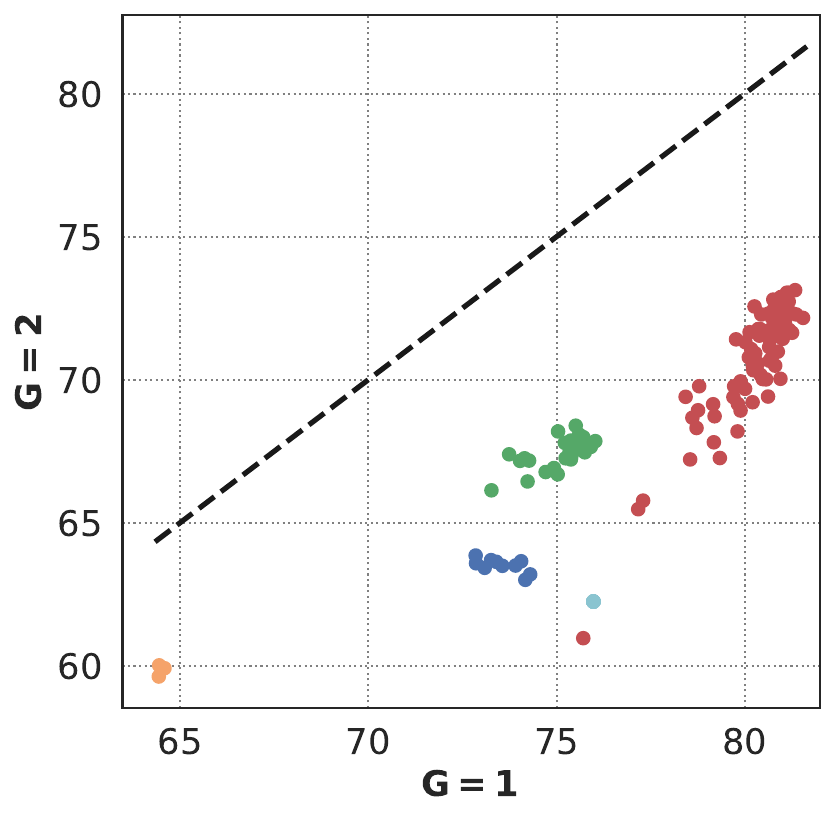}
    \includegraphics[width=0.19\linewidth]{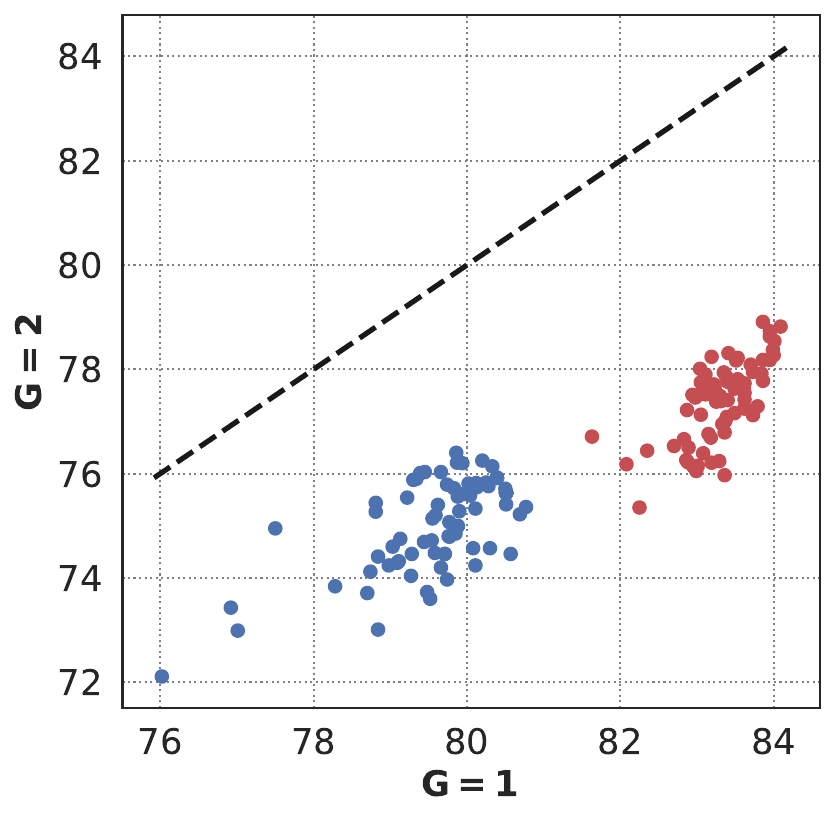}
    
	\begin{minipage}{0.19\linewidth}
    	\small{~~~~~~~~~~~~~~(a) SPOS.}
    \end{minipage}
    \begin{minipage}{0.19\linewidth}
    	\small{~~~~~~~~~~~~(b) FS-NAS.}
    \end{minipage}
    \begin{minipage}{0.19\linewidth}
    	\small{~~~~~~~~~~~~~~~(c) Ours.}
    \end{minipage}
    \begin{minipage}{0.19\linewidth}
    	\small{~~~~~~~~~~~~(d) FS-NAS.}
    \end{minipage}
    \begin{minipage}{0.19\linewidth}
    	\small{~~~~~~~~~~~~~~~(e) Ours.}
    \end{minipage}
    \vspace{-.25cm}
    \caption{Analysis of the performance of subnets with varying the number of channels for supernets. The x-axis shows accuracies of subnets sampled from supernets using full channel dimensions~(\emph{i.e.}, $G$=1), while the y-axis represents those sampled from supernets with reduced channel~(\emph{i.e.}, $G$=2). We measure the rank correlation in terms of Kendall's tau scores~\cite{kendall1938new}, particularly for high-performing subnets (\emph{i.e.}, top 150 subnets). (a-c) We have observed that exploiting multiple supernets enables better preserving the performance ranking. (d-e) We speculate that this is because the performance ranking among subnets from different supernets is likely to be maintained, as the subnets belonging to different supernets do not interfere with each other. Note that we highlight each supernet in a different color. Best viewed in color.}
    \label{fig:fig4}
    \vspace{-.5cm}
\end{figure*}

\noindent \textbf{Channel adjustment.} Exploiting multiple supernets alleviates the interference between subnets remarkably, but at the expense of more computational costs than one-shot NAS methods~\cite{guo2020single,chu2021fairnas,you2020greedynas}. Let us suppose that we have $K$ supernets in total. Then, the total amount of parameters becomes $K$ times more than that of one-shot methods, making it difficult to train supernets simultaneously on a single machine. As a result, existing few-shot methods~\cite{zhao2021few,hu2022generalizing} typically train supernets in a sequential manner, which is time-consuming. A straightforward way to handle this problem is to reduce the number of channels for each supernet, but the reduced channel dimensions could have detrimental effects on NAS as shown in~\cite{liu2018darts,xu2019pc}. To verify this, we compare in Fig.~\ref{fig:fig4} the ranking correlation between two cases: one is with the full channels and the other is with the reduced channels. To this end, we compute Kendall's tau scores~\cite{kendall1938new}, particularly for high-performing subnets (\emph{i.e.}, top 150 subnets). We can see in Fig.~\ref{fig:fig4}(a) that SPOS~\cite{guo2020single} using a single supernet suffers from the ranking inconsistency after reducing the number of channels. On the contrary, we can see in Figs.~\ref{fig:fig4}(b-c) that both FS-NAS~\cite{zhao2021few} and our method better preserve the performance ranking between the high-performing subnets. This suggests that effectively dividing the search space provides the robustness to reducing the channel dimensions. A plausible reason could be that the performance ranking between subnets sampled from different supernets is likely to be maintained (See Figs.~\ref{fig:fig4}(d-e)).

Based on this observation, we propose to reduce the number of channels required for each supernet as follows:
\begin{equation}
	\mathcal{W}_{G}({\mathcal{A}_k}) = \bigcup_{i,j} w_{k}(G,i,j),
\end{equation}
where we denote by $G$ a hyperparamter adjusting the number of channels for each operation, and we define the parameters of the $i$-th operation at the $j$-th layer with reduced channel dimensions as follows:
\begin{equation}
	w_{k}(G,i,j) \in \mathbb{R}^{\frac{C_{\text{out}}(i,j)}{G} \times \frac{C_{\text{in}}(i,j)}{G} \times s(i,j) \times s(i,j)},
\end{equation}
which enables training supernets simultaneously on a single machine. The comparison between our approach and one-shot NAS methods in terms of the total amount of parameters can be represented as follows:
\begin{equation}
	\frac{K\vert \mathcal{W}_{G}(\mathcal{A}_k)\vert}{\vert \mathcal{W}(\mathcal{A})\vert} = \frac{K\vert w_{k}(G,i,j)\vert}{\vert w(i,j)\vert} = \frac{K}{G^2}.
\end{equation}
Note that the number of parameters for our method is comparable to that of one-shot methods by setting the proper value of~$G$~(\emph{e.g.}, $G$=2 if $K$=4).

\begin{figure}[t]
	\small
	\centering
    \includegraphics[width=0.9\linewidth]{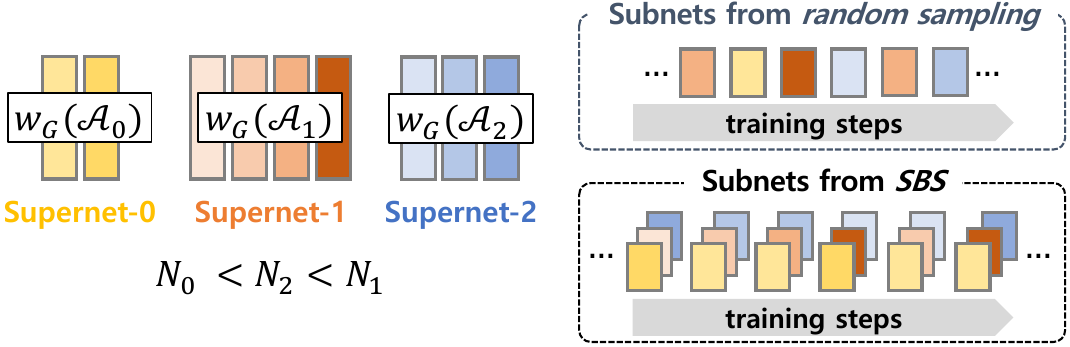}
    \vspace{-.3cm}
    \caption{\textbf{Left:} We visualize three supernets, where the supernet colored in orange has the largest number of subnets $N_1$. \textbf{Right:} Randomly sampling a single subnet from the entire space $\mathcal{A}$~(=$\mathcal{A}_0 \cup \mathcal{A}_1 \cup \mathcal{A}_2$) at each training step causes the training imbalance between supernets. That is, the training process is biased towards sampling subnets from the space of~$\mathcal{A}_1$~(top). Instead, our SBS samples multiple subnets at each training step, where each subnet is sampled from a different supernet (bottom). This allows us to train supernets evenly within a finite number of training steps. Best viewed in color.}
    \label{fig:fig5}
    \vspace{-.6cm}
\end{figure}

\subsection{Training with SBS}
A straightforward way to train supernets is to sample a single subnet randomly from the entire search space~$\mathcal{A}$ at each training step as in~\cite{guo2020single,you2020greedynas,zhao2021few,hu2022generalizing}. However, this might be problematic in that the total number of subnets within each supernet varies significantly~(Fig.~\ref{fig:fig5} (left)). That is, subnets sampled from the entire space are likely to belong to the supernet with the largest number of subnets~(Fig.~\ref{fig:fig5} (right-top)), causing the training imbalance between supernets. To address this, we propose to sample one subnet from each supernet, resulting in a total of $K$ subnets at each training step~(Fig.~\ref{fig:fig5} (right-bottom)). Concretely, we optimize the following objective at each training step:
\begin{equation}
	\sum_{k} \mathcal{L}_{\text{tr}}(a_n; \mathcal{W}_{G}(\mathcal{A}_{k})),
\end{equation}
which aggregates each training loss~$\mathcal{L}_{\text{tr}}$ using the subnet $a_n$ whose parameters are inherited from the corresponding supernet~$\mathcal{W}_G(\mathcal{A}_k)$. This allows us to train all supernets equally during training, avoiding the training imbalance.

\subsection{Searching}
After training supernets, we search the best-performing subnet from the search space. To this end, we estimate the performance of a subnet by inheriting its parameters from a corresponding supernet. Concretely, we can obtain the optimal architecture~$a_\ast$ as follows:
\begin{equation}
	a_\ast = \operatorname*{argmin}_{a_n \in \mathcal{A}} \mathcal{L}_{\text{val}}(a_n; \mathcal{W}_{G}(\mathcal{A}_{m})),
\end{equation}
where $\mathcal{L}_{\text{val}}$ is a validation loss and we denote by $m$ the number of nonlinear functions within the sampled subnet $a_n$. However, traversing all subnets is infeasible, since the total number of subnets~$N$ is typically innumerable (\emph{e.g.}, $6^6\times7^{15}$ in the MobileNet search space~\cite{cai2018proxylessnas,sandler2018mobilenetv2}). Following the common practice~\cite{guo2020single,you2020greedynas,su2021k}, we exploit an evolutionary search algorithm~\cite{guo2020single} as an efficient compromise. Please refer to the supplementary material for more descriptions and a pseudo code for the evolutionary search.


\section{Experiments}
In this section, we describe implementation details and provide quantitative results on standard NAS benchmarks~\cite{krizhevsky2009learning,krizhevsky2012imagenet,chrabaszcz2017downsampled}.

\subsection{Implementation details}
\noindent \textbf{Datasets and search spaces.} We perform experiments on standard NAS benchmarks for image classification: CIFAR10~\cite{krizhevsky2009learning} and ImageNet~\cite{krizhevsky2012imagenet}. CIFAR10 provides 50K training and 10K test samples for 10 object classes, while ImageNet consists of 1.2M training and 50K validation samples for 1K object classes. Following the standard protocol in~\cite{liu2018darts,xu2019pc,guo2020single,hu2020angle}, we split the training set of CIFAR10 in half and use each for training and validation, respectively. For ImageNet, we sample 50K images from the training set to construct a new validation set, and use the original validation set for testing. We adopt NAS201~\cite{dong2020bench} and MobileNet~\cite{cai2018proxylessnas,sandler2018mobilenetv2} search spaces for CIFAR10 and ImageNet, respectively. Specifically, NAS201 is a micro search space using a fixed cell structure, where each cell has five operations with six layers (\emph{i.e.}, edges), resulting in $5^{6}$ architectures. For ImageNet, we use a macro search space consisting of $6^6\times7^{15}$ architectures in total. We use the number of nonlinear functions to split a search space, unless otherwise specified. Since both NAS201 and MobileNet search spaces have candidate operations using ReLU~\cite{krizhevsky2012imagenet} as a nonlinear function, we thus count the number of ReLU functions within a subnet. Note that our splitting criterion is applicable regardless of the types of activation functions. Please see the supplementary material for a detailed description of the candidate operations.

\noindent \textbf{Training and evaluation.} We follow the common practice~\cite{zhao2021few,hu2022generalizing} for training supernets and retraining optimal architectures. Specifically, we train supernets for 200 epochs with a batch size of $1,024$. We adopt a SGD optimizer with an initial learning rate of 0.12, a momentum of 0.9, and a weight decay of 4$e$-5. The learning rate is adjusted by a cosine annealing strategy without restart. The number of supernets $K$ is set to 3 and 6 on NAS201~\cite{krizhevsky2009learning} and ImageNet~\cite{krizhevsky2012imagenet}, respectively. We set $G$ to 2 for all experiments, unless otherwise specified. After applying the evolutionary search algorithm~\cite{guo2020single}, we train the chosen architectures for 450 epochs with a batch size of $1,024$. We use a RMSProp optimizer with an initial learning rate and a weight decay of 0.064 and 1$e$-5, respectively. The learning rate decays by a factor of 0.97 per 2.4 epochs. For evaluation, we compute top-1 and top-5 classification accuracies, and report average scores using 3 different seeds for all experiments. All experiments are performed with 8 NVIDIA A5000 GPUs.

\begin{table}[t]
	\small
	\centering
	\caption{Comparison of the rank correlation in terms of Kendall's tau scores~\cite{kendall1938new} on the test set of CIFAR10~\cite{krizhevsky2009learning}. Our approach outperforms one- and few-shot NAS methods by a large margin, while adopting three supernets with half channel dimensions (\emph{i.e.}, $G$=2). Note that it requires fewer parameters than one-shot NAS methods. Numbers in bold are the best performance and underlined ones are the second best. Params: the number of parameters required for supernets.}
	\vspace{-.2cm}
	\addtolength{\tabcolsep}{1pt}
	\label{tab:kendall}
	\begin{tabular}{l c c c}
		\specialrule{.1em}{.05em}{.05em}
		\multicolumn{1}{c}{\multirow{2}{*}{Method}} & \multirow{2}{*}{$K$} & Params & \multirow{2}{*}{Kendall's $\tau$}
		\\
		& & (M) &
		\\
		\hline
        \rowcolor[gray]{.9} \multicolumn{4}{c}{\textit{One}-shot NAS}
		\\
		\hline
		SPOS~\cite{guo2020single} & 1 & ~~\underline{1.7} & 0.554
		\\
		AngleNet~\cite{hu2020angle} & 1 & ~~\underline{1.7} & 0.575
		\\
		\hline
        \rowcolor[gray]{.9} \multicolumn{4}{c}{\textit{Few}-shot NAS}
		\\
		\hline
		FS-NAS~\cite{zhao2021few} & 5 & ~~8.4 & 0.653
		\\
		GM-NAS~\cite{hu2022generalizing} & 8 & 13.6 & 0.656
		\\
		$K$-shot NAS~\cite{su2021k} & 8 & 13.6 & 0.626
		\\
		Ours & & &
		\\
		~~FLOPs & 3 & ~~\textbf{1.3} & {0.711}
		\\
		~~\# of linear regions & 3 & ~~\textbf{1.3} & \underline{0.712}
		\\
		~~Feature isotropy & 3 & ~~\textbf{1.3} & {0.693}
		\\
		~~\# of nonlinear functions & 3 & ~~\textbf{1.3} & \textbf{0.735}
		\\
		\specialrule{.1em}{.05em}{.05em}	
	\end{tabular}
	\vspace{-.5cm}
\end{table}

\begin{table}[t]
	\small
	\centering
	\caption{Test accuracies of searched architectures on NAS201~\cite{dong2020bench}. Different from current few-shot methods using multiple supernets with full channel dimensions, our method uses three supernets with half channel dimensions (\emph{i.e.}, $G$=2). C10, C100, and IN represent CIFAR10, CIFAR100~\cite{krizhevsky2009learning}, and ImageNet-16-120~\cite{chrabaszcz2017downsampled}, respectively.}
	\vspace{-.2cm}
	\addtolength{\tabcolsep}{-2pt}
	\label{tab:sota_c10}
	\begin{tabular}{l c c c}
		\specialrule{.1em}{.05em}{.05em}
		\multicolumn{1}{c}{Method} & C10 & C100 & IN
		\\ 
		\hline
        \rowcolor[gray]{.9} \multicolumn{4}{c}{\textit{One}-shot NAS}
		\\
		\hline
		DARTS~\cite{liu2018darts} & 54.30 & 15.61 & 16.32
		\\
		PC-DARTS~\cite{xu2019pc} & 93.41 & 67.48 & 41.31
		\\
		SPOS~\cite{guo2020single} & 93.67 & 69.83 & 44.71
		\\
		AngleNet~\cite{hu2020angle} & 94.01 & 72.96 & 45.83
		\\
		AGNAS~\cite{sun2022agnas} & 94.05 & 72.41 & 45.98
		\\
		\hline
        \rowcolor[gray]{.9} \multicolumn{4}{c}{\textit{Few}-shot NAS}
		\\
		\hline
		FS-NAS~\cite{zhao2021few} & 89.11 & 58.69 & 33.85
		\\
		GM-NAS~\cite{hu2022generalizing} & 92.70 & 68.81 & 43.47
		\\
		$K$-shot NAS~\cite{su2021k} & \underline{94.19} & \textbf{73.45} & \underline{46.53}
		\\
		Ours & \textbf{94.30} & \underline{73.20} & \textbf{46.60}
		\\
		\hline
		{Upper bound} & 94.37 & 73.51 & 47.31
		\\
		\specialrule{.1em}{.05em}{.05em}	
	\end{tabular}
	\vspace{-.7cm}
\end{table}

\subsection{Results}
\noindent \textbf{NAS201.} We compare in Table~\ref{tab:kendall} our approach with one- and few-shot NAS methods on the test set of CIFAR10~\cite{krizhevsky2009learning}. Specifically, we measure the Kendall rank correlation~\cite{kendall1938new} between stand-alone accuracies of subnets provided by NAS201~\cite{dong2020bench} and estimated ones from supernets. From this table, we can see that few-shot NAS approaches~\cite{zhao2021few,su2021k} achieve better results than one-shot baselines~\cite{guo2020single,hu2020angle}, suggesting that reducing the extent of weight sharing improves the rank consistency remarkably. We can also see that our approach, regardless of the types of zero-cost proxies, outperforms other few-shot methods by a large margin, demonstrating its effectiveness. This is particularly significant in that we require fewer parameters for supernets than one-shot NAS methods, while the total number of parameters for current few-shot methods increases with the number of supernets. Among the zero-cost proxies, using the number of nonlinear functions gives the best performance, suggesting that it groups subnets effectively. We show in Table~\ref{tab:sota_c10} top-1 test accuracies of chosen architectures. We can see that our approach provides the high-performing architectures on each dataset of NAS201. This validates once again that our splitting criterion is simple yet effective.

\noindent \textbf{MobileNet.} We report in Table~\ref{tab:sota_in} top-1 and top-5 accuracies of our architectures chosen from the MobileNet search space~\cite{cai2018proxylessnas,sandler2018mobilenetv2}. To this end, we perform the evolutionary search~\cite{guo2020single} using FLOPs as a hardware constraint. We can see from this table that our method with a constraint of 530M FLOPs provides better results than FS-NAS~\cite{zhao2021few} and GM-NAS~\cite{hu2022generalizing} in terms of test accuracy, FLOPs, and the number of parameters. This is remarkable in that our method exploits six supernets with reduced channel dimensions to alleviate the computational cost, while FS-NAS and GM-NAS adopt five and six supernets with full channel dimensions, respectively. We can also see that our architecture searched with a constraint of 600M FLOPs achieves the highest test accuracy. This suggests the importance of effectively dividing the search space to improve the search performance.

\begin{table}[t]
	\small
    \centering
    \caption{Quantitative results of searched architectures on ImageNet~\cite{deng2009imagenet}. We use two constraints in terms of FLOPs for the evolutionary search algorithm~\cite{guo2020single}. FS-NAS~\cite{zhao2021few} and GM-NAS~\cite{hu2022generalizing} exploit five and six supernets with full channel dimensions, respectively, while our method adopts six supernets with half channel dimensions (\emph{i.e.}, $G$=2). Params: the number of network parameters for the chosen architecture.}
	\addtolength{\tabcolsep}{-4pt}
    \label{tab:sota_in}
	\vspace{-.2cm}
	\begin{tabular}{l c c c c}
	\specialrule{.1em}{.05em}{.05em}
	\multicolumn{1}{c}{\multirow{2}{*}{Method}} & \multicolumn{2}{c}{Acc. (\%)} & FLOPs & Params
	\\
	\cline{2-3}
    & Top-1 & Top-5 & (M) & (M)
	\\
    \hline
    \rowcolor[gray]{.9} \multicolumn{5}{c}{\textit{One}-shot NAS}
    \\
    \hline
    GAEA~\scriptsize{\cite{li2020geometry}} & 76.0 & 92.7 & - & 5.6
    \\
    SPOS~\scriptsize{\cite{guo2020single}} & 74.7 & - & 328 & 3.4
    \\
    ProxylessNAS~\scriptsize{\cite{cai2018proxylessnas}} & 75.1 & 92.5 & 465 & 7.1
    \\
    AngleNet~\scriptsize{\cite{hu2020angle}} & 76.1 & - & 470 & -
    \\
    Shapley-NAS~\scriptsize{\cite{xiao2022shapley}} & 76.1 & - & 582 & 5.4
    \\
    PC-DARTS~\scriptsize{\cite{xu2019pc}} & 75.8 & 92.7 & 597 & 5.3
    \\
    DrNAS~\scriptsize{\cite{chen2020drnas}} & 76.3 & 92.9 & 604 & 5.7
    \\
    ISTA-NAS~\scriptsize{\cite{yang2020ista}} & 76.0 & 92.9 & 638 & 5.7
    \\
    \hline
    \rowcolor[gray]{.9} \multicolumn{5}{c}{\textit{Few}-shot NAS}
	\\
    \hline
    FS-NAS~\scriptsize{\cite{zhao2021few}} & 75.9 & - & 521 & 4.9
    \\
    GM-NAS~\scriptsize{\cite{hu2022generalizing}} & {76.6} & \underline{93.0} & 530 & 4.9
    \\
    Ours ($\leq$530M) &  \underline{76.7} & \textbf{93.2} & 516 & 4.8
	\\
    Ours ($\leq$600M) & \textbf{76.9} & \textbf{93.2} & 544 & 4.9
	\\
	\specialrule{.1em}{.05em}{.05em}	
    \end{tabular}
    \vspace{-.3cm}
\end{table}

\begin{table}[t]
	\small
	\centering
	\caption{Analysis of different values of $K$ on ImageNet~\cite{deng2009imagenet}. We report the computational cost in terms of GPU days along with the top-1 accuracies of the searched architectures. Specifically, we denote by splitting and training the amount of time to split a search space and train supernets, respectively. We run all experiments on the same machine with 8 NVIDIA A5000 GPUs.}
	\addtolength{\tabcolsep}{-2pt}
	\label{tab:diff_k_g}
	\vspace{-.2cm}
	\begin{tabular}{l c c c c}
		\specialrule{.1em}{.05em}{.05em}
		\multicolumn{1}{c}{\multirow{2}{*}{Method}} & \multicolumn{2}{c}{Cost} & Top-1 & FLOPs
		\\
		\cline{2-3}
		& splitting & training & Acc. (\%) & (M)
		\\
		\hline
		FS-NAS~\scriptsize{\cite{zhao2021few}} & - & 23.1 & 75.9 & 521
		\\
		GM-NAS~\scriptsize{\cite{hu2022generalizing}} & 17.0 & 81.0 & 76.6 & 530
		\\
		Ours: $K$=4~\&~$G$=2 & - & 17.3 & 76.5 & 527
		\\
		Ours: $K$=6~\&~$G$=2 & - & 23.3 & 76.7 & 516
		\\
		Ours: $K$=8~\&~$G$=2 & - & 34.1 & 76.8 & 522
		\\
		\specialrule{.1em}{.05em}{.05em}	
	\end{tabular}
	\vspace{-.6cm}
\end{table}

\subsection{Discussion}
\noindent \textbf{Analysis of the numbers of supernets.} We provide in Table~\ref{tab:diff_k_g} results for the selected architectures with varying the values of $K$ on ImageNet~\cite{krizhevsky2012imagenet}. For comparison, we report the results of FS-NAS~\cite{zhao2021few} and GM-NAS~\cite{hu2022generalizing}. We can see from this table three things: (1) Similar to FS-NAS, our splitting criterion induces a negligible overhead. On the contrary, GM-NAS incurs a lot of computational cost particularly for splitting the search space. In addition, the training cost of GM-NAS is approximately 3.5 times higher than our method using the same number of supernets (\emph{i.e.}, $K$=6). This is because GM-NAS trains supernets sequentially due to the large memory requirements. (2) We can see from the last three rows that using more supernets provides better results in terms of test accuracy at the cost of increasing the training time. For example, our method using eight supernets gives an accuracy gain of 0.2\% over GM-NAS with a much lower training cost. (3) We can save the training cost remarkably by reducing the channel dimensions for supernets (\emph{i.e.}, $G$=2). For example, setting $G$ to 2 for training six supernets induces the training time comparable to FS-NAS, while outperforming GM-NAS in terms of test accuracies.


\section{Conclusion}
We have introduced a novel few-shot NAS method that exploits the number of nonlinear functions to split a search space into a set of disjoint subspaces. Our splitting criterion is simple yet effective, making subnets from different supernets have distinct characteristics in terms of test accuracy, FLOPs, and the number of parameters. Based on our novel observation that effectively dividing the space provides the robustness to reducing the number of channels, we have proposed to adjust the channel dimensions required for supernets to alleviate the computational cost, allowing us to train supernets on a single machine. We have also presented a SBS technique, which samples multiple subnets belonging to different supernets at each training step, to train all supernets evenly within a limited number of training steps. Finally, we have demonstrated the effectiveness of our approach on standard NAS benchmarks.


\section{Acknowledgments}
This work was supported in part by the NRF and IITP grants funded by the Korea government (MSIT) (No.2023R1A2C2004306, No.RS-2022-00143524, Development of Fundamental Technology and Integrated Solution for Next-Generation Automatic Artificial Intelligence System), the KIST Institutional Program (Project No.2E31051-21-203), and the Yonsei Signature Research Cluster Program of 2024 (2024-22-0161).
\bibliography{aaai25}

\clearpage


\section*{Supplementary material}

# Efficient Few-Shot Neural Architecture Search by Counting the Number of Nonlinear Functions
# Supplement

**Youngmin Oh[1], Hyunju Lee[1], and Bumsub Ham[1,2]***

[1]Yonsei University   [2]Korea Institute of Science and Technology (KIST)
{youngmin.oh, hyunjulee, bumsub.ham}@yonsei.ac.kr

In this supplementary material, we first describe more details of the evolutionary search (Guo et al. 2020), along with pseudo code (Sec. 1). We then provide detailed descriptions of the candidate operations for NAS201 (Dong and Yang 2020) and MobileNet (Sandler et al. 2018; Cai, Zhu, and Han 2019) (Sec. 2). We also present more discussions including results on NAS201 and our limitation (Sec. 3). Finally, we visualize the chosen architectures (Sec. 4).

## 1 Evolutionary search

We summarize in Algorithm 1 the overall process of evolutionary search. Following SPOS (Guo et al. 2020), we use the default hyperparameters. To be specific, we set the numbers of population $P$ and generations $T$ to 50 and 20, respectively, evaluating a total of 1K subnets. The initial generation $\mathcal{S}_0$ is defined as a set of $P$ subnets sampled from a search space $\mathcal{A}$ randomly. To obtain a new generation $\mathcal{S}_t$, we first choose the top ten subnets (*i.e.*, $B$=10) in terms of validation accuracy, whose FLOPs are lower than the hardware constraint $C_{\text{FLOP}}$, from the previous generation $\mathcal{S}_{t-1}$. Then, we perform mutation and crossover functions with the chosen subnets. The mutation function produces a new subnet by changing each operation of a randomly selected subnet with a probability, $p_m$, of 0.1. On the other hand, the crossover function generates a new subnet by selecting one operation from two randomly selected subnets at each layer.

## 2 Search spaces

**NAS201.** NAS201 (Dong and Yang 2020) provides a micro search space designed for discovering optimal cell structures, where the final network architecture is determined by stacking the searched cell repeatedly. Each cell is a densely-connected directed acyclic graph consisting of four nodes and six edges (*i.e.*, layers). There are four main paths from the input node to the output node (See Fig. A(a)). Each layer has five candidate operations, resulting in a total of $15,625$ possible cells (*i.e.*, architectures). We show in Table A each operation along with its number of nonlinear functions. From this table, we can see that the first three operations do not contain any nonlinear functions. Specifically, 'none'

---

---

Algorithm 1: Pseudo code of evolutionary search.

1: **Input:** the number of population $P$, the number of generations $T$, the number of selections $B$, a hardware constraint $C_{\text{FLOP}}$ in terms of FLOPs, the mutation probability $p_m$
2: **Output:** the optimal architecture $a_*$
3: // Set the number of offspring, *i.e.*, $M$ to $P/2$
4: $M = P/2$
5: // Randomly sample $P$ subnets from a search space $\mathcal{A}$
6: $\mathcal{S}_0 = \{a_n\}_{n=1}^{P}$ from $\mathcal{A}$
7: **for** $t = 1$ to $T$ **do**
8:   // Evaluate a set of subnets $\mathcal{S}_{t-1}$ on a validation set
9:   $\text{Acc}_{t-1} = evaluate(\mathcal{S}_{t-1})$
10:  // Get the top-$B$ subnets with FLOPs less than $C_{\text{FLOP}}$
11:  Top-$B = get\_top(\text{Acc}_{t-1}, B, C_{\text{FLOP}})$
12:  // Mutate the top-$B$ subnets to generate $M$ new subnets
13:  $\mathcal{S}^{\text{m}} = mutation(\text{Top-}B, M, p_m)$
14:  // Crossover the top-$B$ subnets to generate $M$ new subnets
15:  $\mathcal{S}^{\text{c}} = crossover(\text{Top-}B, M)$
16:  // Get a new generation $\mathcal{S}_t$
17:  $\mathcal{S}_t = \mathcal{S}^{\text{m}} \cup \mathcal{S}^{\text{c}}$
18: **end for**
19: // Get the optimal subnet $a_*$
20: $\text{Acc}_T = evaluate(\mathcal{S}_T)$
21: $a_* = get\_top(\text{Acc}_T, 1, C_{\text{FLOP}})$

---

indicates removing the corresponding layer, 'skip_connect' represents an identity mapping, and 'avg_pool_3x3' is a 3x3 average pooling layer. We can also see from the last two rows that 'conv_$s$x$s$', which represents a stack of a ReLU (Krizhevsky, Sutskever, and Hinton 2012) activation, a $s$x$s$ convolutional layer, and a batch normalization (Ioffe and Szegedy 2015) layer, has one nonlinear function. To count the valid number of nonlinear functions within a cell (See Fig. A(b)), we define the following rules: (1) Ignore a path containing the 'none' operation, since the input and output nodes are disconnected. (2) Accumulate the number of nonlinear functions, if layers are connected in series. (3) Select a path with the maximum number of nonlinear functions. According to the rules, there are four possible values of $k$ (*i.e.*, the number of nonlinear functions) on NAS201: 0, 1, 2, and 3. We divide $15,625$ cells into three groups (*i.e.*, subspaces) as shown in Table B, where we assign two subspace (*i.e.*, $\mathcal{A}_0$ and $\mathcal{A}_1$) to one supernet. Note that subnets from the two subspaces $\mathcal{A}_0$ and $\mathcal{A}_1$ do not interfere with

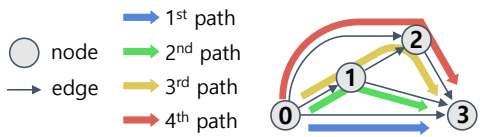

(a) Four paths from the input node to the output node within a cell.

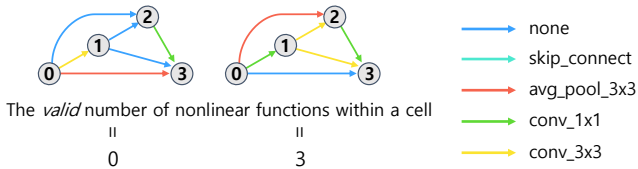

(b) Counting the valid number of nonlinear functions within a cell.

Figure A: Visualization of a cell structure on NAS201 (Dong and Yang 2020). (a) We show four possible paths from the input node to the output node. (b) We describe how to count the valid number of nonlinear functions within a cell. Specifically, we ignore a path containing the 'none' operation, and choose a path with the maximum number of nonlinear functions to represent the valid number of nonlinear functions within a cell. Best viewed in color.

Table A: Candidate operations for the NAS201 (Dong and Yang 2020) search space. We provide the number of nonlinear functions for each operation. Please see text for more details.

| candidate operations | # of nonlinear functions |
|---|---|
| none | 0 |
| skip_connect | 0 |
| avg_pool_3x3 | 0 |
| conv_1x1 | 1 |
| conv_3x3 | 1 |

each other, since the 'conv_sxs' operations of subnets from the subspace $\mathcal{A}_0$ are not trained at all.

**MobileNet.** ProxylessNAS (Cai, Zhu, and Han 2019) introduces a macro search space on ImageNet (Krizhevsky, Sutskever, and Hinton 2012) using building blocks of MobileNetV2 (Sandler et al. 2018). There are a total of 21 searchable layers, where the numbers of reduction and normal layers are 6 and 15, respectively. We show in Table C seven candidate operations at each normal layer, where ID indicates an identity mapping and MB$u$_K$v$ represents an inverted bottleneck (IB) block with an expansion ratio of $u$ and a kernel size of $v$. Each IB block has two ReLU (Krizhevsky, Sutskever, and Hinton 2012) activations, while ID does not contain nonlinear functions. The reduction layers contain 6 candidate operations only, excluding ID. Note that the MobileNet search space does not have the 'none' operation. Consequently, the total number of network architectures becomes $6^6 \times 7^{15}$, and there are a total of 16 possible values of $k$, where the minimum and maximum values are 12 and 42, respectively. However, employing 16 supernets would be computationally demanding even with reducing the channels as in Eq. (7) of the main paper. We thus force a particular supernet to cover multiple subspaces, especially those with low values of $k$. This is because we have observed that test accuracies of subnets tend to increase in proportion to the

Table B: Subspaces covered by each supernet for the NAS201 (Dong and Yang 2020) search space. Please see text for more details.

| $K$ | Supernet Index | Subspace |
|---|---|---|
| 3 | 1 | $\mathcal{A}_3$ |
| | 2 | $\mathcal{A}_2$ |
| | 3 | $\mathcal{A}_0, \mathcal{A}_1$ |

Table C: Candidate operations for the MobileNet (Cai, Zhu, and Han 2019; Sandler et al. 2018) search space. We provide the number of nonlinear functions for each operation. Please see text for more details.

| candidate operations | # of nonlinear functions |
|---|---|
| ID | 0 |
| MB3_K3 | 2 |
| MB3_K5 | 2 |
| MB3_K7 | 2 |
| MB6_K3 | 2 |
| MB6_K5 | 2 |
| MB6_K7 | 2 |

value of $k$. For example, we can see from Fig. 2(e) in the main paper that the first supernet (*i.e.*, the supernet covering $\mathcal{A}_3$ in Table B) shows the highest median accuracy.

## 3 More discussions

**Analysis of the proxy-based criteria.** To demonstrate the efficiency of our criterion, we compare in Table E time requirements w.r.t 15,625 subnets on NAS201 and MobileNet spaces. For the MobileNet space, we randomly sample the subnets. We can see from this table that the total time required to compute LR and ISO for 15,625 subnets is much higher than the time required for our criterion. In addition, we can also see that the runtime for LR and ISO increases when applied to the MobileNet space. This is because computing LR and ISO for a subnet depends on both the size of the input image and the number of network parameters. By contrast, our criterion is input- and parameter-agnostic, making it more efficient. We'd like to note that saving a single forward pass is critical for both splitting the search space and evaluating subnets, since the number of subnets is innumerable (*e.g.*, the MobileNet space has $6^6 \times 7^{15}$ subnets).

To further demonstrate the effectiveness of our criterion, we report in Fig. B training losses averaged across supernets. To be specific, we compare five splitting criteria: (1) FS-NAS (Zhao et al. 2021) dividing the space randomly; (2) 'Ours w/ FP' that uses the number of FLOPs of a subnet (Ning et al. 2021; Li et al. 2023) as a splitting criterion; (3) 'Ours w/ LR' that uses the number of linear regions divided by ReLU activations (Mellor et al. 2021) as a splitting criterion; (4) 'Ours w/ ISO' that uses a feature isotropy (Lee and Ham 2024) as a splitting criterion; (5) 'Ours' using the number of nonlinear functions. We can see from the figure that our approach to using the number of nonlinear functions always achieves lower values than other criteria regardless of $G$, confirming that our method better groups subnets so that they are less likely to interfere with each other. Note

Table D: Subspaces covered by each supernet in accordance with the number of supernets for the MobileNet (Cai, Zhu, and Han 2019; Sandler et al. 2018) search space. Please see text for more details.

| $K$ | Supernet Index | Subspace |
|---|---|---|
| 4 | 1 | $\mathcal{A}_{42}$ |
|  | 2 | $\mathcal{A}_{40}$ |
|  | 3 | $\mathcal{A}_{38}$ |
|  | 4 | $\mathcal{A}_{12}, \mathcal{A}_{14}, \mathcal{A}_{16}, \mathcal{A}_{18}, \mathcal{A}_{20}, \mathcal{A}_{22},$ $\mathcal{A}_{24}, \mathcal{A}_{26}, \mathcal{A}_{28}, \mathcal{A}_{30}, \mathcal{A}_{32}, \mathcal{A}_{34}, \mathcal{A}_{36}$ |
| 6 | 1 | $\mathcal{A}_{42}$ |
|  | 2 | $\mathcal{A}_{40}$ |
|  | 3 | $\mathcal{A}_{38}$ |
|  | 4 | $\mathcal{A}_{36}$ |
|  | 5 | $\mathcal{A}_{34}$ |
|  | 6 | $\mathcal{A}_{12}, \mathcal{A}_{14}, \mathcal{A}_{16}, \mathcal{A}_{18}, \mathcal{A}_{20}, \mathcal{A}_{22},$ $\mathcal{A}_{24}, \mathcal{A}_{26}, \mathcal{A}_{28}, \mathcal{A}_{30}, \mathcal{A}_{32}$ |
| 8 | 1 | $\mathcal{A}_{42}$ |
|  | 2 | $\mathcal{A}_{40}$ |
|  | 3 | $\mathcal{A}_{38}$ |
|  | 4 | $\mathcal{A}_{36}$ |
|  | 5 | $\mathcal{A}_{34}$ |
|  | 6 | $\mathcal{A}_{32}$ |
|  | 7 | $\mathcal{A}_{30}$ |
|  | 8 | $\mathcal{A}_{12}, \mathcal{A}_{14}, \mathcal{A}_{16}, \mathcal{A}_{18}, \mathcal{A}_{20}, \mathcal{A}_{22},$ $\mathcal{A}_{24}, \mathcal{A}_{26}, \mathcal{A}_{28}$ |

Table E: Runtime comparison of different splitting criteria. LR and ISO indicate the number of linear regions (Mellor et al. 2021) and the feature isotropy (Lee and Ham 2024), respectively.

| Time (m) | NAS201 | MobileNet |
|---|---|---|
| Ours | 0 | 0 |
| Ours w/ LR | 7 | 46 |
| Ours w/ ISO | 3 | 29 |

that the average training loss is likely to increase when there are conflicts between subnets.

**Comparison of Kendall's tau scores.** Table F provides a detailed analysis of how our splitting criterion improves the performance on CIFAR10 (Krizhevsky, Hinton et al. 2009) of NAS201 (Dong and Yang 2020). Specifically, we measure the rank correlation for each subspace in terms of Kendall's tau scores (Kendall 1938). The table shows that our method shows significant improvements for every subspace consistently, validating once again the importance of effectively dividing the search space. A plausible reason is that our splitting criterion makes subnets from the same subspace have similar structures (See Fig. 3 in the main paper), enabling better mitigating the conflicts between subnets than the random splitting (Zhao et al. 2021). In particular, it is worth noting that our method provides the largest gain over the baseline using a single supernet (Guo et al. 2020) for the first subspace consisting of high-performing subnets (See the first column in Fig. 2(e) of the main paper). For a fair comparison, we also provide in Table G the rank correlation for five subspaces divided by the random splitting criterion (Zhao et al. 2021). From this table, we can see that our

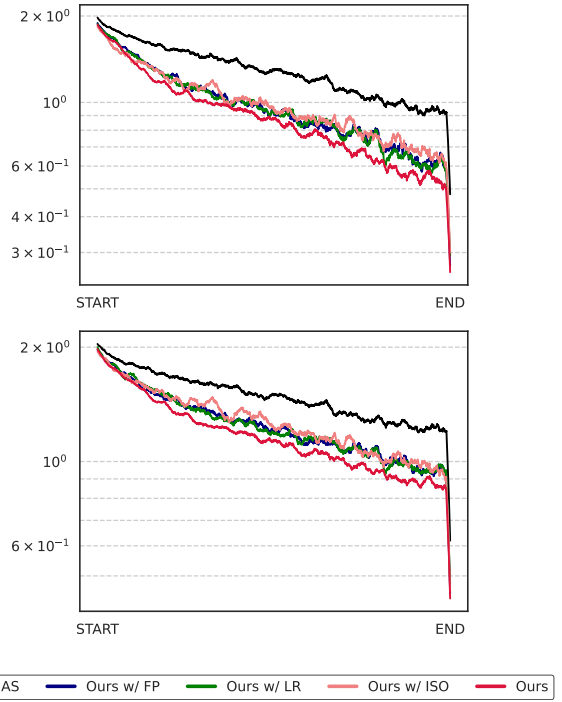

Figure B: Comparison of the average training losses over supernets on CIFAR10 (Krizhevsky, Hinton et al. 2009). We set $G$ to 1 (top) and 2 (bottom).

Table F: Comparison of Kendall's tau scores (Kendall 1938) on CIFAR10 (Krizhevsky, Hinton et al. 2009) of NAS201 (Dong and Yang 2020). We report the scores for the entire space along with three different subspaces. The results of SPOS (Guo et al. 2020) and FS-NAS (Zhao et al. 2021) are obtained from our implementation. $1st$, $2nd$, and $3rd$ indicate the first, second, and third subspaces, respectively.

| Method | Kendall's $\tau$ | | | |
|---|---|---|---|---|
|  | $\mathcal{A}$ | $1st$ | $2nd$ | $3rd$ |
| SPOS (Guo et al. 2020) | 0.577 | 0.429 | 0.543 | 0.668 |
| FS-NAS (Zhao et al. 2021) | 0.580 | 0.464 | 0.470 | 0.619 |
| Ours: $K$=3 & $G$=2 | **0.735** | **0.648** | **0.662** | **0.747** |

method outperforms SPOS and FS-NAS by a large margin for each subspace. This confirms once again that our splitting criterion better splits the search space, mitigating conflicts between subnets at training time.

**Analysis of SBS.** To show the effectiveness of SBS, we compare in Table H results of our approach without and with SBS on NAS201 (Dong and Yang 2020). We can see from the table that our approach without SBS degrades the ranking performance in terms of Kendall's tau scores (Kendall 1938) on CIFAR10 (Krizhevsky, Hinton et al. 2009). A plausible reason is that randomly sampling a subnet from an entire search space at training time is biased towards training a supernet covering the largest number of subnets. To further verify this, we compare in Table I test accuracies of chosen architectures. The table shows that our approach with SBS outperforms the one without SBS remarkably on each dataset of NAS201, demonstrating the importance of SBS.

**Analysis of training losses.** We visualize in Fig. C train-

Table G: Comparison of Kendall's tau scores (Kendall 1938) on CIFAR10 (Krizhevsky, Hinton et al. 2009). We report the scores for the entire space along with five different subspaces divided by FS-NAS (Zhao et al. 2021). The results of SPOS (Guo et al. 2020) and FS-NAS are obtained from our implementation. $1st$, $2nd$, $3rd$, $4th$, and $5th$ indicate the first, second, third, fourth, and fifth subspaces, respectively.

| Method | Kendall's $\tau$ | | | | | |
|---|---|---|---|---|---|---|
| | $\mathcal{A}$ | $1st$ | $2nd$ | $3rd$ | $4th$ | $5th$ |
| SPOS (Guo et al. 2020) | 0.577 | 0.658 | 0.628 | 0.482 | 0.444 | 0.663 |
| FS-NAS (Zhao et al. 2021) | 0.580 | 0.673 | 0.651 | 0.550 | 0.530 | 0.620 |
| Ours: $K$=3 & $G$=2 | **0.735** | **0.822** | **0.732** | **0.709** | **0.670** | **0.674** |

Table H: Comparison of the rank correlation in terms of Kendall's tau scores (Kendall 1938) on CIFAR10 (Krizhevsky, Hinton et al. 2009). Params: the number of network parameters required for supernets.

| Method | # of supernets | Params (M) | Kendall's $\tau$ |
|---|---|---|---|
| Ours w/o SBS | 3 | 1.3 | 0.458 |
| Ours | 3 | 1.3 | 0.735 |

ing curves on CIFAR10 (Krizhevsky, Hinton et al. 2009) of NAS201 (Dong and Yang 2020) with setting $G$ to 1 (top) and 2 (bottom). Specifically, we show in Fig. C(a) the average loss over supernets at each training step, while Figs. C(b-d) show a training curve for each supernet individually. For comparison, we provide the results of SPOS (Guo et al. 2020). For SPOS, we show in Fig. C(a) a training loss at each training step, and plot in Figs. C(b-d) the one, only if the sampled subnet belongs to the corresponding supernet. From this figure, we have the following observations: (1) Our method without SBS shows higher training losses than the one with SBS. This is because randomly sampling a single subnet at each training step results in the training imbalance between supernets. For example, since the supernet in Fig. C(d) consists of approximately nine times more subnets than the one in Fig. C(b), our method without SBS shows much higher losses in Fig. C(b). This indicates the training process is biased towards sampling subnets belong to the supernet in Fig. C(d), hindering training the supernet in Fig. C(b). (2) Our method with SBS avoids this problem, always achieving lower losses than other methods regardless of the value of $G$. This suggests that our splitting criterion results in subspaces, where subnets from the same supernet are less likely to interfere with each other during training. Note that the conflicts between subnets hinder the training process, raising the training losses.

**Limitation.** Our approach to using the number of nonlinear functions for few-shot NAS reduces the computational overhead remarkably. It however requires training a few supernets, which is still more demanding than one-shot NAS approaches. Nonetheless, we believe that our observation, which reveals that properly dividing the search space enables maintaining the performance ranking between subnets even after reducing the channel dimension required for supernets, would provide new insights into designing efficient

Table I: Test accuracies of chosen architectures on NAS201 (Dong and Yang 2020).

| Method | CIFAR10 | CIFAR100 | IN-16-120 |
|---|---|---|---|
| Ours w/o SBS | 91.91 | 67.44 | 39.15 |
| Ours | 94.30 | 73.20 | 46.60 |

few-shot NAS methods. In addition, combining several zero-cost proxies to better group subnets would be a promising direction, even if it increases the computational overhead.

## 4 Visualization of chosen architectures

We show in Figs. D and E our architectures searched from the NAS201 (Dong and Yang 2020) and MobileNet (Cai, Zhu, and Han 2019; Sandler et al. 2018) search spaces, respectively. Specifically, Fig. D shows the cell structure searched on CIFAR10 (Krizhevsky, Hinton et al. 2009), while Fig. E shows two architectures searched with hardware constraints of 530M (top) and 600M (bottom) FLOPs on ImageNet (Krizhevsky, Sutskever, and Hinton 2012).

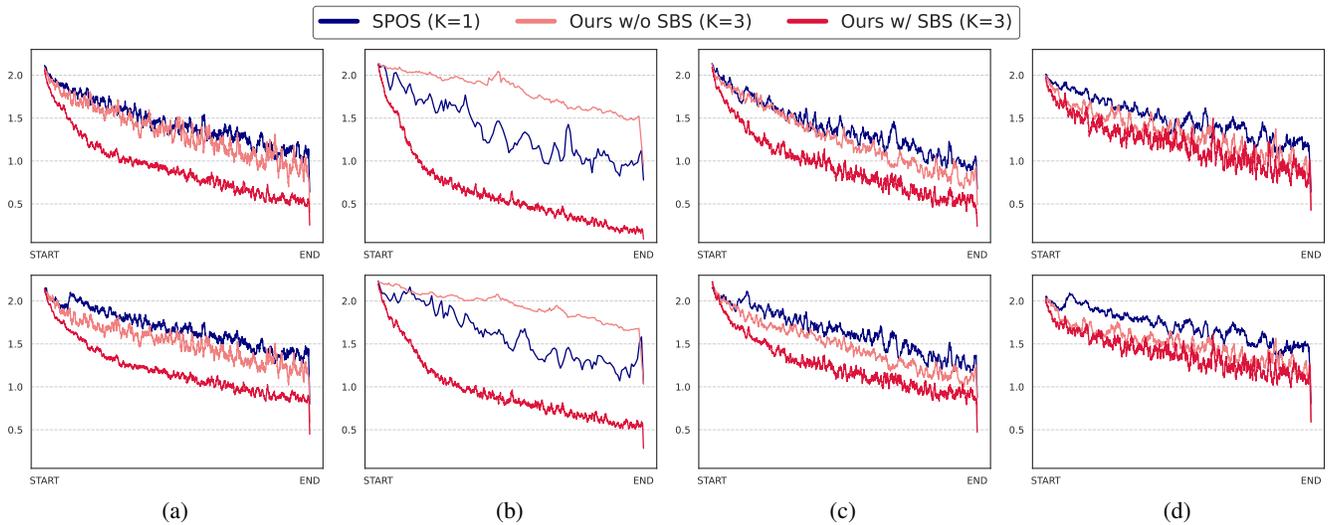

Figure C: Comparison of training curves. We set the value of $G$ to 1 (top) and 2 (bottom). (a) We show the average loss over supernets at each training step. (b-d) We plot a training curve for each supernet individually. The second column shows that our approach without SBS produces much higher losses. This is because subnets are likely to belong to the supernet with the largest number of subnets if we sample a single subnet randomly. Our SBS avoids this problem, enabling achieving lower losses than other methods consistently. Note that the supernet with the largest number of subnets is shown in (d). Best viewed in color.

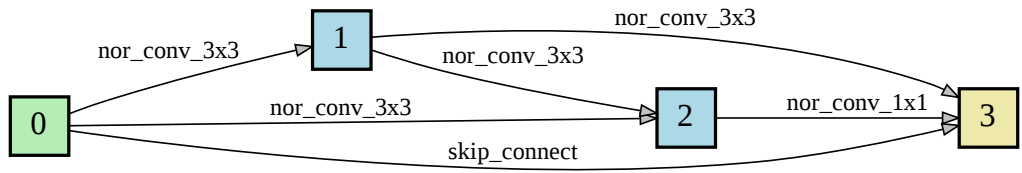

Figure D: Visualization of the chosen cell structure on CIFAR10 (Krizhevsky, Hinton et al. 2009) for the NAS201 (Dong and Yang 2020) search space. We set $K$ and $G$ to 3 and 2, respectively.

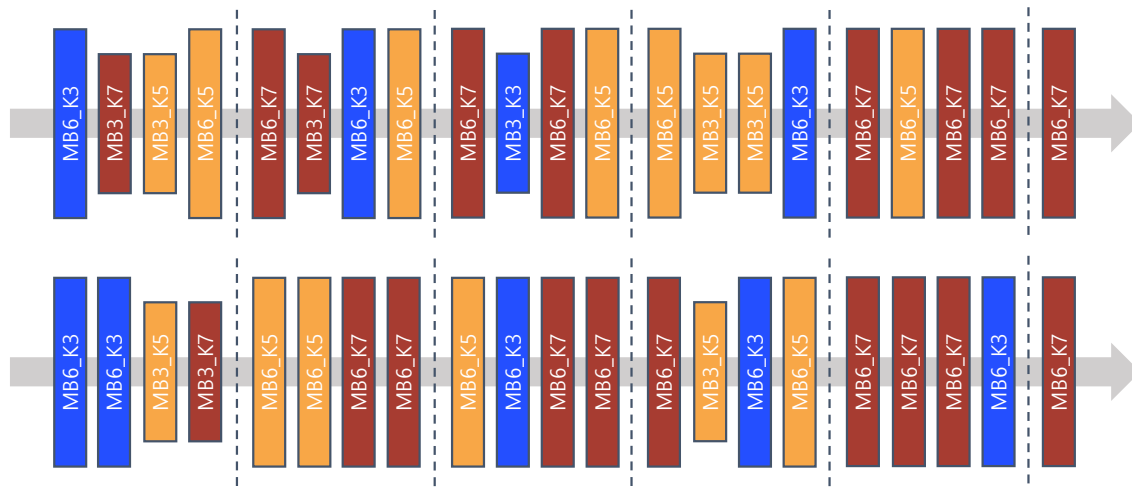

Figure E: Visualization of chosen architectures on ImageNet (Krizhevsky, Sutskever, and Hinton 2012) for the MobileNet (Sandler et al. 2018; Cai, Zhu, and Han 2019) search space. We set $K$ and $G$ to 6 and 2, respectively. For evolutionary search (Guo et al. 2020), we adopt two hardware constraints in terms of FLOPs: 530M (top) and 600M (bottom).



\end{document}